\documentclass{article}

    \usepackage[final, nonatbib]{neurips_2022}

\usepackage[numbers,sort&compress]{natbib}
\usepackage[utf8]{inputenc} 
\usepackage[T1]{fontenc}    
\usepackage{hyperref}       
\usepackage{url}            
\usepackage{booktabs}       
\usepackage{amsfonts,amsmath,amssymb, amsthm}       
\usepackage{nicefrac}       
\usepackage{microtype}      
\usepackage{xcolor}         
\usepackage{graphicx}
\usepackage{times}
\usepackage{latexsym}
\usepackage{dsfont}
\usepackage{multirow}
\usepackage{comment}
\usepackage{placeins}
\usepackage{wrapfig}

\usepackage[font=small,labelfont=bf]{caption}

\usepackage{stmaryrd}

\title{How to talk so AI will learn: \\ Instructions, descriptions, and autonomy} 

\author{
    Theodore R. Sumers \\
    Computer Science \\
    Princeton University \\
    \texttt{sumers@princeton.edu}\\
    \And Robert D. Hawkins  \\
  Princeton Neuroscience Institute \\
  Princeton University \\ 
  \texttt{rdhawkins@princeton.edu}
  \And Mark K. Ho  \\ 
  Computer Science \\
  Princeton University \\ 
  \texttt{mho@princeton.edu}
  \AND
  Thomas L. Griffiths \\
  Computer Science, Psychology \\
  Princeton University  \\
  \texttt{tomg@princeton.edu}
  \And Dylan Hadfield-Menell  \\ 
 EECS, CSAIL \\
  MIT  \\
  \texttt{dhm@csail.mit.edu}
}

\begin{document}

\maketitle

\begin{abstract}
From the earliest years of our lives, humans use language to express our beliefs and desires. Being able to talk to artificial agents about our preferences would thus fulfill a central goal of value alignment.
Yet today, we lack computational models explaining such language use.
To address this challenge, we formalize learning from language in a contextual bandit setting and ask how a human might communicate preferences over behaviors. 
We study two distinct types of language: \emph{instructions}, which provide information about the desired policy, and \emph{descriptions}, which provide information about the reward function. 
We show that the agent's degree of autonomy determines which form of language is optimal: instructions are better in low-autonomy settings, but descriptions are better when the agent will need to act independently. 
We then define a pragmatic listener agent that robustly infers the speaker's reward function by reasoning about \emph{how} the speaker expresses themselves.
We validate our models with a behavioral experiment, demonstrating that (1) our speaker model predicts human behavior, and (2) our pragmatic listener successfully recovers humans' reward functions. Finally, we show that this form of social learning can integrate with and reduce regret in traditional reinforcement learning. 
We hope these insights facilitate a shift from developing agents that \emph{obey} language to agents that \emph{learn} from it.

\end{abstract}

\section{Introduction}
As artificial agents proliferate in society, aligning them with human values is increasingly important~\cite{Gabriel2020alignment, russell2019humancompatible, amodei2016concrete}. But how can we build machines that understand what we want? Prior work has highlighted the difficulty of specifying our desires via numerical reward functions~\cite{amodei2016concrete, hadfield2017inverse, knox2021reward}. Here, we explore language as a means to communicate them. While most previous work on language input to AI systems focuses on \emph{instructions}~\cite{winograd1972understanding, Tellex2011, chen2011learning, macglashan2015grounding, fu_2019_goals, Bahdanau_2019, sharma2022correcting, jeon2020reward, milli2017should, maclin_1994, kuhlman_2004, goyal2019using, luketina2019survey, Tellex2020}, we study instructions alongside more abstract, \emph{descriptive} language~\cite{branavan2012learning, narasimhan2018grounding, zhong2019rtfm, hanjie2021grounding, sumers2021learning, lin2022inferring}. We examine how humans communicate about rewards, and formalize learning from this input. 

To consider how humans communicate about reward functions, imagine taking up mushroom foraging. How would you learn the rewards associated with different fungi (i.e. which are delicious and which are deadly)? In such a setting, learning from direct experience~\cite{sutton2018reinforcement} is risky; most humans would seek to learn \emph{socially} instead.
So how might we learn reward functions from others?
Prior work in reinforcement learning (RL) has examined a number of social learning strategies, including passive \emph{inverse reinforcement learning} (observe an expert pick mushrooms, then infer their reward function~\cite{ng2000algorithms, abbeel2004apprenticeship}) or active \emph{preference learning} (offer an expert pairs of mushrooms, observe which one they eat, and infer their reward function~\cite{markant2014better, christiano2017deep, basu2018learning}).

However, few humans would rely on such indirect data if they had access to a cooperative teacher \cite{velez2021learning,gweon2021inferential,wang2020mathematical, shafto_pedagogical}. 
For example, an expert guiding a foraging trip might \emph{demonstrate}~\cite{ho2016showing, Ho2021communication} or verbally \emph{instruct}~\cite{Tellex2011} the learner to pick certain mushrooms, licensing stronger inferences.
While such pedagogical actions have been useful for guiding RL agents \cite{goyal2019using,luketina2019survey,fu_2019_goals,Tellex2020}, natural language affords richer, under-explored forms of teaching. 
For example, an expert teaching a seminar might instead \emph{describe} how to recognize edible or toxic mushrooms based on their features, thus providing highly generalized information.

\begin{figure*}[t]
    \centering
    \includegraphics[width=13.9cm]{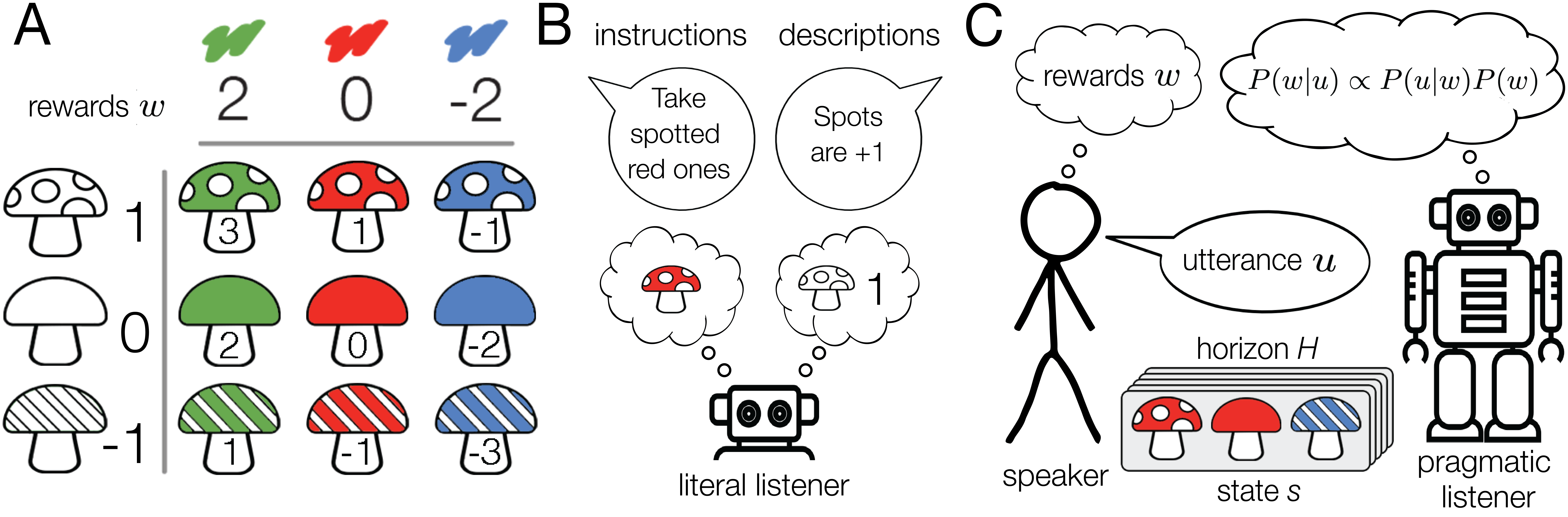} 
    \caption{We formalize learning from language in a contextual bandit setting. \textbf{A}: Linear rewards associated with features determine whether actions (mushrooms) are high or low reward. \textbf{B}: We consider two forms of language. \emph{Instructions} tell the listener to take a specific action, while \emph{descriptions} provide information about underlying reward function. A \emph{literal} listener interprets utterances according to these fixed semantics. \textbf{C}: In this work, we introduce a speaker model suggesting that humans use instructions and descriptions adaptively to maximize a literal listener's expected rewards over some \emph{horizon}, which formalizes the agent's degree of autonomy. We then define a \emph{pragmatic} listener which uses Bayesian inference to recover the speaker's latent reward function.}
    \label{fig:linear_bandits_setup}
\end{figure*}

To formalize this process, we present a model of learning from language in a contextual bandit setting (Fig~\ref{fig:linear_bandits_setup}). We propose a speaker model that chooses utterances to maximize the listener's expected rewards over some task \emph{horizon}. The horizon quantifies notions of autonomy described in previous work~\cite{milli2017should, jeon2020reward, lin2022inferring}. If the horizon is short, the agent is closely supervised; at longer horizons, they are expected to act more independently. We then analyze \emph{instructions} (which provide a partial policy) and \emph{descriptions} (which provide partial information about the reward function). We show that instructions are optimal at short horizons, while descriptions are optimal at longer ones.

Second, we consider how a listener might learn from such a speaker. We define a pragmatic listener that infers the speaker's latent reward function based on their utterances. While prior work suggests pragmatic learning can be vulnerable to model mis-specification~\cite{milli2020literal}, we show that jointly inferring the speaker's horizon and reward function can mitigate this risk. 

Finally, we conduct an online behavioral experiment which shows our models support strong reward inference. We integrate this social inference with traditional RL and find that it accelerates learning and reduces regret. Overall, our results suggest that descriptive language and pragmatic inference are powerful mechanisms for value alignment and learning.\footnote{Code and data are available at \url{https://github.com/tsumers/how-to-talk}.}

\section{Related work}

Classic RL assumes that the reward function is given to the agent~\cite{sutton2018reinforcement}. However, in practice, it is difficult to specify a reward function to obtain desired behavior~\cite{amodei2016concrete}, motivating \emph{learning} the reward function from social input.\footnote{Another approach bypasses learning the reward function and allows the human to provide a reward signal directly~\cite{Knox2009Interactively, macglashan2017interactive}. This is challenging, however, as people do not typically provide classical reinforcement~\cite{Ho2019punishment}.} 
Most social learning methods assume the expert is simply acting optimally~\cite{ng2000algorithms, abbeel2004apprenticeship, ziebart2008maximum}, but recent \textit{pragmatic} methods~\cite{ho2016showing, HadfieldMenell2016cooperative, fisac2020pragmatic} instead assume the expert is actively teaching.

\textbf{Learning reward functions from observed actions.} When the desired behavior is known (but the reward function is not), inverse reinforcement learning~\cite[IRL, ][]{ng2000algorithms, abbeel2004apprenticeship,  ramachandran2007bayesian, ziebart2008maximum, milli2017should, jeon2020reward, Tulli2022maxtenthumans} can be used to infer an expert's reward function from their actions. However, such approaches face fundamental issues with identifiability: observed behavior can often be explained by multiple reward functions. One solution allows the agent to actively query the human~\cite{christiano2017deep, basu2018learning, Christian2015active, sadigh2017active, palan2019learning}. An alternative is to make a stronger assumption: that the human is actively teaching~\citep[i.e. behaving \emph{pedagogically}, ][]{ho2016showing,HadfieldMenell2016cooperative,fisac2020pragmatic}. We next review work on learning rewards from language, then return to these methods.

\textbf{Learning reward functions from instructions.}
Instructions use language to communicate specific actions or goals~\cite{winograd1972understanding, Tellex2011, chen2011learning}. Prior work has used instructions to shape reward functions~\cite{maclin_1994, kuhlman_2004, goyal2019using} or learn a language-conditioned reward model~\cite{Bahdanau_2019, macglashan2015grounding, fu_2019_goals, sharma2022correcting}. An alternative approach uses instructions for value alignment. This paradigm assumes instructions reflect underlying preferences and uses them to infer this latent reward function~\cite{jeon2020reward, milli2017should}.
Our work extends this to incorporate other forms of language: we ask when a human would use an instruction (vs.~other forms of language), and---given an instruction (vs.~other forms of language)---what an agent should infer about the speaker's reward function.

\textbf{Learning reward functions from descriptions.} Rather than expressing specific goals, reward-descriptive language encodes abstract information about preferences or the world. 
The education literature suggests such rich feedback is crucial for human learning~\cite{shute_2008, Lipnevich_education}. 
However, a relatively smaller body of work uses it for RL: by learning reward functions directly from existing bodies of text~\cite{branavan2012learning, narasimhan2018grounding, zhong2019rtfm, hanjie2021grounding} or interactive, free-form language input~\cite{sumers2021learning, lin2022inferring}. 
Our work provides a formal model of such language in order to compare it with more typically studied instructions.
Other related lines of work use language which describes agent \emph{behaviors}. This language, whether externally provided~\cite{coreyes2018guiding, sharma2021skill, wong2021leveraging, andreas2018latentlang, chen2021ask, nguyen2021interactive, jiang2019language, ren2022leveraging} or internally generated~\cite{colas2020language, cideron2020higher, mu2022improving, tam2022semantic}, is typically used to structure the task representation or guide exploration. In contrast, we study language describing task-relevant properties of the environment.

\textbf{Learning reward functions from pedagogy.}
The preceding algorithms all assume that training examples (whether demonstrations or language) are generated by a human that is \textit{indifferent to the learning process}. 
Recent work has begun to challenge this assumption by considering \emph{pedagogical} settings~\cite{shafto_pedagogical, wang2020mathematical, ho2016showing, Ho2021communication}.
In particular, the Rational Speech Act framework~\cite{frank2012pragmatic, goodman_2016} builds on classic Gricean theory~\cite{grice1975logic} to formulate optimal communication in terms of recursive Bayesian inference and decision-making.
These ideas have been applied to a variety of language tasks including reference games~\cite{monroe2017colors,andreas2016pragmaticreference}, captioning~\cite{nie2020pragmaticcaptions}, and instruction following~\cite{fried2018pragmatics}. 
Developing analogues of linguistic pragmatics in reinforcement learning --- i.e., algorithms that assume data are \textit{intentionally designed to be informative} --- is an active area of research~\cite{fisac2020pragmatic, milli2020literal,ho2016showing,HadfieldMenell2016cooperative}. In particular, inverse reward design (IRD,~\cite{hadfield2017inverse}) uses pragmatic inference on numerical reward functions: rather than take the provided reward function literally,  IRD quantifies uncertainty over the function to mitigate alignment risk. Similarly, we apply pragmatic inference to reward-related language. 
Applying IRD principles to reward-related language offers two primary benefits over its non-linguistic formulation. First, language is \emph{accessible} for humans. While traditional IRD is useful when offline training RL agents, inferring rewards from language would be broadly useful for real-world, real-time interactions with non-experts. Second, language can address \emph{future settings}: speakers can refer to actions or features which are not physically present. Speakers can thus provide information about rare or hazardous possibilities (e.g. poisonous mushrooms) which lie outside the listener's experience. We next describe a model for such language use.

\section{Background}
\label{sec_rsa_background}
In this work, we consider the problem of \emph{learning from language} in a \emph{contextual bandit} setting. This section introduces models of language use from cognitive science, then describes contextual bandits.

\textbf{The Rational Speech Acts  framework (RSA)}.
Our theoretical approach is based on the Rational Speech Acts framework (RSA,~\cite{goodman_2016, frank2012pragmatic, yoon2016polite, Kao2014nonliteral, kao2014metaphor}), a computational model of language understanding. 
RSA uses theory-of-mind reasoning to explain how human listeners derive meaning from words. 
It begins by defining a \textit{literal listener} $L_0$ which considers only the conventional (e.g., dictionary) definitions of words. These conventions are represented by a ``lexicon'' function $\mathcal{L}$ which maps from utterances to world states $w$.\footnote{The lexicon can be seen as the \emph{grounding} of language~\cite{harnad1990symbol, mooney2008learning}. Most RSA models, including this work, assume that groundings are mutually known. However, methods have been developed to learn them~\cite{sumers2021learning, lin2022inferring} or allow for uncertainty~\cite{degen2020redundancy,hawkins2022partners}, and we view integrating the problem of grounding as important future work.}
RSA then assumes that a \textit{speaker} $S_1$ makes a noisy-optimal selection from a finite set of utterances, choosing $u$ according to some utility function $U(\cdot)$.
Finally, a \textit{pragmatic listener} $L_1$ infers the intended meaning by inverting the speaker model:
\begin{align}
P_{L_0}(w \mid u) & \propto  \mathcal{L}(u, w) \label{eq_lexicon}\\
P_{S_1}(u \mid w) & \propto  \exp\{\beta_{S_1} \cdot U(u,w)\} \label{eq_s1_likelihood}\\
P_{L_1}(w \mid u) & \propto  P_{S_1}(u \mid w)P(w)
\label{eq_l1_inference}
\end{align}
In Eq.~\ref{eq_s1_likelihood}, $\beta_{S_1} > 0$ is an inverse  temperature parameter. RSA models typically assume the speaker's objective is \emph{informativeness}. Formally, the speaker tries to reduce the listener's information-theoretic uncertainty~\cite{cover1999elements} over the true state of the world $w^*$: $U(u, w^*) = \ln P_{L_0}(w^* \mid u)$. 
Explicitly modeling the speaker as making a rational choice from a set of possible utterances allows the pragmatic listener to enrich the literal meaning of the utterance, inferring additional information about the world.

To take an intuitive example, we consider the classic phenomenon of \emph{scalar implicature}~\cite{levinson1983pragmatics, levinson2000presumptive, goodman2013knowledge}. If Alice announces ``I ate some of the cookies,'' a typical listener Bob understands this as ``some \emph{but not all}.'' RSA explains this as follows. First, the word ``some'' \emph{literally} maps to any world state where Alice ate more than zero cookies (e.g. 1, 2, 3... or all), while ``all'' maps only to the world state where she ate all of them (Eq.~\ref{eq_lexicon}). Then, $S_1$ Alice  is presumed to choose a maximally-informative utterance (i.e. one that specifies the world state as precisely as possible; Eq.~\ref{eq_s1_likelihood}). Finally, $L_1$ Bob reasons as follows: if Alice had eaten all of the cookies, saying ``I ate \emph{all} of the cookies'' would have been more precise and thus preferred. 
Because she chose \emph{not} to say this, he can \emph{infer} that her statement means ``some \emph{but not all}'' (Eq.~\ref{eq_l1_inference}). In this work, we define a pragmatic $L_1$ listener which uses RSA to infer the speaker's reward function. Crucially, such inferences are based on theory-of-mind and thus vulnerable to \emph{misspecification}~\cite{milli2020literal}: if $L_1$ Bob's model of $S_1$ Alice is wrong, his inferences will be wrong. We show how a carefully specified speaker model, incorporating both instructions and descriptions, allows our pragmatic listener to robustly infer a speaker's reward function.

\textbf{Contextual bandits.}
Contextual bandits can be seen as one-step Markov Decision Processes~\cite{puterman1994markov, sutton2018reinforcement}, thus isolating the problem of \emph{learning} a reward function from the problem of \emph{planning} sequential actions to maximize it. As such, they are a popular testbed for RL algorithms~\cite{bandit_algorithms2020, amin2017repeated, riquelme2018deep, chu2011contextual, Abbasi2011bandits}. Formally, we define a set of $A$ possible actions. Actions are associated with a binary feature vector $\phi: A \rightarrow \{0, 1\}^K$ (e.g., a mushroom may be green (or not), or striped (or not)). Following other work in IRL~\cite{ng2000algorithms, ziebart2008maximum, hadfield2017inverse}, we assume rewards are a linear function of these features (e.g., green mushrooms tend to be tasty):
\begin{equation}
    R(a, w) = w^\top\phi(a)
\label{eq_linear_rewards}
\end{equation}
so $w$ is a vector that defines the value of each feature (see Fig.~\ref{fig:linear_bandits_setup}A). 
Each task consists of a sequence of $H$ i.i.d. states. We refer to $H$ as the \emph{horizon.}  
At each time step $t<H$, the agent is presented with a state $s_t$ consisting of a subset of possible actions: $s_t \subseteq A$ (e.g., a patch containing a set of mushrooms). 
They choose an action $a \in s_t$ according to their policy, $\pi_L: S \rightarrow \Delta (A)$.

\section{Formalizing learning from language in contextual bandits}
\label{section_reward_design}

While bandits are typically considered as an \emph{individual} learning problem, we instead ask how an agent should learn \emph{socially} from a cooperative, knowledgeable partner. 
We formalize this by introducing a second agent: a speaker who knows the true rewards $w$ and the initial state $s_0$, and produces an utterance $u$ (Fig.~\ref{fig:linear_bandits_setup}C). This utterance may affect the listener's policy; with a slight abuse of notation, we denote this updated policy as $\pi_L( a \mid u,s)$. The listener then uses this policy to choose actions. 
The horizon $H$ determines how many actions the listener will perform under this updated policy. 
Intuitively, $H=1$ represents minimal autonomy (or maximal supervision, i.e. guided foraging), whereas $H \to \infty$ is maximal autonomy (or minimal supervision, i.e. teaching the listener to forage independently). 
We first assume $H$ is known to both listener and speaker, then relax this assumption. 
This framework exposes two interrelated problems.
First, what should a helpful speaker say? 
And second, how should the listener update their policy in light of this information? 

\begin{figure*}[t]
    \centering
    \includegraphics[width=13.9cm]{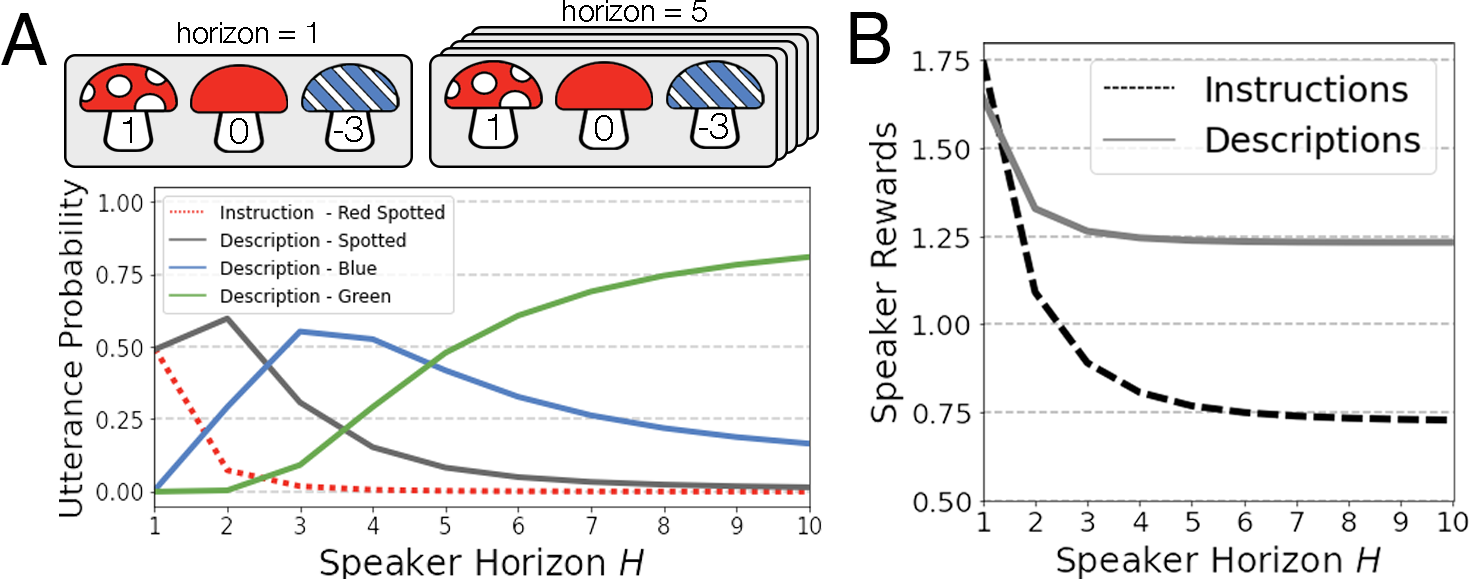} 
    \caption{The \emph{horizon} formalizes the agent's degree of autonomy and strongly affects a reward-maximizing speaker's choice of utterance. \textbf{A}: Top: as the horizon lengthens, the agent will act independently in more unknown future states. Bottom: simulating a reward-maximizing speaker. At short horizons, the speaker is focused on the spotted red mushroom. As the horizon lengthens, they are less biased towards this particular state and provide more generally useful information about the reward function. \textbf{B}: Speaker rewards (Eq.~\ref{eq_speaker_combined_utility}, averaged over all 84 start states) for a speaker with access to only instructions or only descriptions. Instructions outperform descriptions at short horizons, but descriptions offer much stronger generalization.
    }
    \label{fig:fig2-speaker-horizon-instruction-descriptions}
\end{figure*}

\subsection{Speaker model} 
We apply the RSA framework (Section~\ref{sec_rsa_background}) to a RL setting by introducing a new speaker objective: instead of Gricean informativeness~\cite{grice1975logic, goodman_2016} we assume that speakers seek to \emph{maximize the listener's rewards}~\cite{sumers2021extending}. Conceptually, this changes RSA's scope from epistemic to decision-theoretic. Instead of reasoning about how utterances affect the listener's \emph{beliefs}, such a speaker reasons about how they affect the listener's \emph{policy}.
When the state is known, we define the \emph{present} utility of an utterance as the expected reward from using the updated policy to choose an action: 
\begin{equation}
    U_\text{Present}(u \mid s, w) = \sum_{a \in s} \pi_{L}(a \mid u, s) R(a, w)
\label{eq_local_rewards}
\end{equation}
However, we are interested in settings where agents may need to act autonomously~\cite{lin2022inferring, jeon2020reward, sumers2021learning}. Thus, we must consider how well the policy generalizes to unknown, future states. 
The \emph{future} utility of an utterance with respect to a distribution over states $P(s)$ can be written as:
\begin{equation}
    U_\text{Future}(u \mid w) = \sum_{s \in S} U_\text{Present}(u \mid s, w)P(s)
\label{eq_expected_rewards}
\end{equation}
Because states are i.i.d. in the bandit setting, a speaker optimizing for a horizon $H$ can be defined as a linear combination of Eqs.~\ref{eq_local_rewards} and~\ref{eq_expected_rewards}:
\begin{equation}
    U_{S_1}(u \mid w, s, H) = \frac{U_\text{Present} + (H-1)U_\text{Future}}{H} = \frac{1}{H}U_\text{Present} + (1 - \frac{1}{H})U_\text{Future}
\label{eq_speaker_combined_utility}
\end{equation}
where $H=1$ reduces to Eq.~\ref{eq_local_rewards} and as $H\to\infty$ reduces to Eq.~\ref{eq_expected_rewards}. At short horizons, speakers are biased towards producing utterances that optimize the listener's policy in the present state. As the horizon lengthens, the speaker expects the listener to behave more autonomously. Generalization becomes more important, and this bias is reduced.

\subsection{Formalizing instructions}
We now consider how utterances affect the listener's policy. {\em Instructions} map to specific actions or trajectories~\cite{Tellex2011, jeon2020reward}; in our work, ``instruction'' utterances correspond to the nine actions (Fig~\ref{fig:linear_bandits_setup}B). Given an instruction, a literal listener executes the corresponding action. If the action is not available, the listener acts randomly:
\vspace{-4mm}
\begin{equation}
\pi_{L_0}(a \mid u_\text{instruction}, s) =  
\begin{cases}
    0 & \text{if } a \notin s \\
    \delta_{\llbracket u \rrbracket (a)} & \text{if } \llbracket u \rrbracket \in s\\
    \frac{1}{\mid s \mid} & \text{otherwise}
\end{cases}
\label{eq_literal_listener_instruction}
\end{equation}
where $\delta_{\llbracket u \rrbracket(a)}$ represents the meaning of $u$, evaluating to one when utterance $u$ grounds to $a$ and zero otherwise. An instruction is a \emph{partial policy}: it designates an action to take in a subset of states.

\subsection{Formalizing descriptions}
\label{section_descriptions}
Rather than mapping to a specific action, descriptions provide information about the reward function~\cite{ sumers2021learning, lin2022inferring, Tulli2022maxtenthumans}. Following~\cite{sumers2021extending}, we model descriptions as providing the reward of a single feature, similar to feature queries~\cite{basu2018learning}. Descriptions are thus a tuple: a one-hot binary feature vector and a scalar value, $\langle\mathds{1}_K, \mathbb{R}\rangle$. These are messages like $\langle$Blue, -2$\rangle$. In this work, we consider the set of $6 \text{ features} \times 5$ values in $[-2, -1, 0, 1, 2]$, yielding 30 descriptive utterances. 
Formally, $L_0$ ``rules out'' inconsistent hypotheses about reward weights $w$:
\begin{equation}
    L_0(w \mid u_\text{description}) \propto \delta_{\llbracket u \rrbracket (w)} P(w)
\label{eq_literal_listener_belief_update}
\end{equation} 
where $\delta_{\llbracket u \rrbracket(w)}$ represents the meaning of $u$, evaluating to one when $u$ is true of $w$ and zero otherwise. In this work, we assume $P(w)$ is uniform and there is no correlation between weights.
The listener then marginalizes over possible reward functions to choose an action:
\begin{equation}
    \pi_{L_0}(a \mid u_\text{description}, s) \propto \exp\{\beta_{L_0} \cdot \sum_w R(a, w)L_0(w \mid u))\}
\label{eq_listener_policy}
\end{equation}
where $\beta_{L_0}$ is again an inverse temperature parameter.\footnote{Due to the recursive structure of speaker and listener, simultaneously varying $\beta_{S_1}$ and $\beta_{L_0}$ introduces identifiability issues. We therefore fix $\beta_{L_0}=3$ throughout this work, and only vary $\beta_{S_1}$. This implies that all speakers address the same listener, but do so with different degrees of optimality.}

\subsection{Comparing instructions and descriptions}
\label{subsec_instructions_vs_descriptions}
Prior work suggests that humans use a mix of instructions and descriptions~\cite{sumers2021learning, lin2022inferring}. What modulates this---when should a rational speaker prefer instructions over descriptions? To explore the effects of horizon on utterance utility, we simulate a nearly-optimal speaker ($\beta_{S_1} = 10$). Fig.~\ref{fig:linear_bandits_setup}A shows our bandit setting. We assume the listener begins with a uniform prior over reward weights throughout, and use states consisting of three unique actions (giving 84 possible states). 

Fig~\ref{fig:fig2-speaker-horizon-instruction-descriptions}A shows how the speaker's choice of utterance varies as the horizon lengthens. At short horizons, the speaker optimizes for present rewards (Eq.~\ref{eq_local_rewards}) and chooses instructions and descriptions that target the ``Spotted Red'' action. At longer horizons, future rewards (Eq.~\ref{eq_expected_rewards}) play a larger role, and the speaker blends the two objectives (Eq.~\ref{eq_speaker_combined_utility}) by describing the highly negative blue feature. Finally, at sufficiently long horizons, future rewards dominate and it settles on describing the green feature---which is irrelevant to the start state, but the most important feature for generalization. 
To quantify how rational speakers should use instructions and descriptions, we repeat the task for all 84 start states using horizons ranging 1-10 and different utterance sets. Fig~\ref{fig:fig2-speaker-horizon-instruction-descriptions}B plots rewards for speakers with access to \emph{only} instructions \emph{or} descriptions, illustrating why this shift from instructions to descriptions occurs. Instructions outperform descriptions at short horizons (achieving the theoretical maximum average reward of 1.75); as the horizon lengthens, however, descriptions generalize better.\footnote{Because our states consist of only three actions, it is often possible to find a description that uniquely identifies the best action. This allows descriptions to perform nearly as well as instructions at $H=1$. However, as the number of available actions increases, instructions become increasingly advantageous; see Appendix~\ref{sec_appdx_pragmatic_simulations}.} Overall, as the horizon lengthens, speakers with access to both instructions and descriptions choose descriptions exclusively, producing highly generalizable information (see Appendix~\ref{sec_appdx_pragmatic_simulations}).


\begin{figure}[t]
    \centering
    \includegraphics[width=13.9cm]{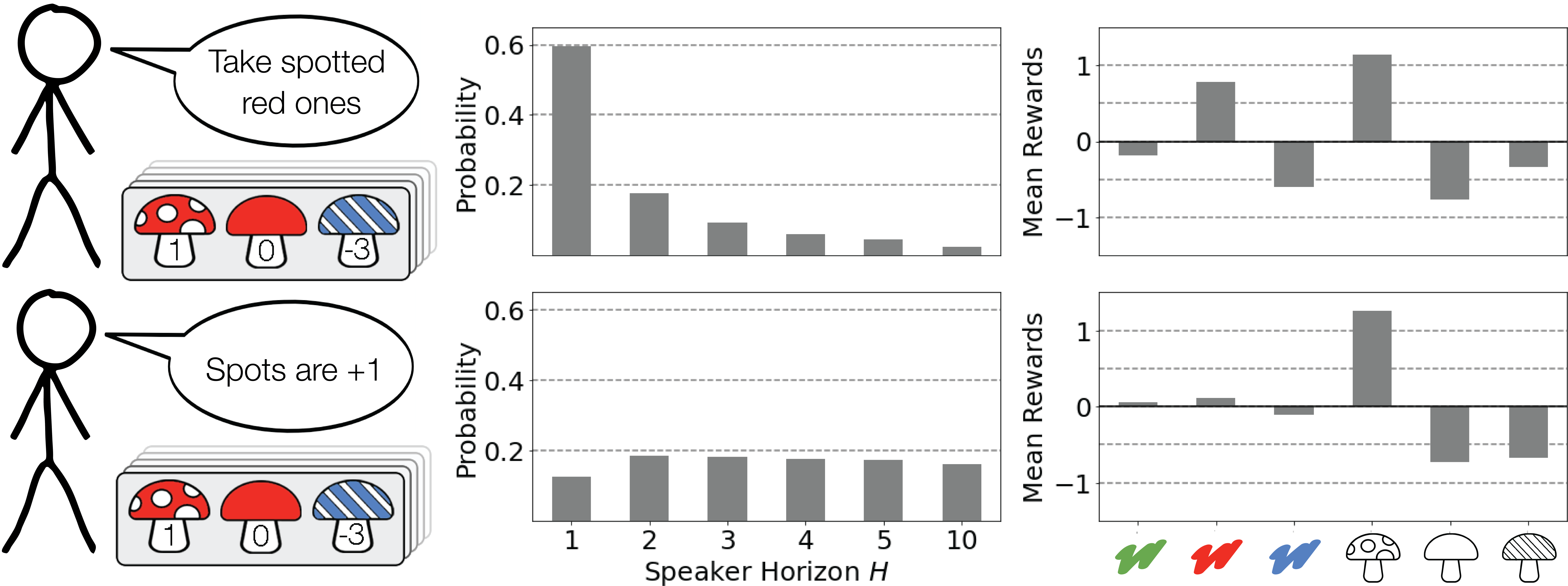}
    \caption{Posteriors from latent-horizon pragmatic inference (Eq.~\ref{eq_joint_pragmatic_listener}) for instructions and descriptions. Left: both utterances suggest the listener should take the spotted red mushroom in this state. Center: an instruction (top) suggests the speaker is short-horizon, while a description (bottom) suggests they are long-horizon. Right: this leads to different inference about the speaker's reward function (see Appendix~\ref{sec_appdx_pragmatic_simulations} for full posteriors).}
    \label{fig:pragmatics}
\end{figure}

\subsection{Learning from utterances: a pragmatic listener}
\label{section_ird}
We now ask how the listener should \emph{learn} from the speaker's utterance. Following RSA (Section~\ref{sec_rsa_background}), a pragmatic listener $L_1$ inverts the speaker model to infer their latent reward function. This approach is closely related to inverse reward design~\cite{hadfield2017inverse}: because the speaker maximizes rewards (Eq.~\ref{eq_speaker_combined_utility}), our pragmatic listener effectively treats utterances as \emph{proxy rewards}~\cite{singh2009rewards} specified in language.

\textbf{Known horizon.} If the speaker's horizon $H$ is known, we can write a $L_1$ listener as:
\begin{equation}
    L_1(w \mid s, u, H) \propto S_1(u \mid w, s, H)P(w)
\label{eq_fixed_horizon_pragmatics}
\end{equation}
Given an instruction, $L_1$ infers the reward weights that would make such an instruction optimal~\cite{milli2017should, jeon2020reward}; given a description, $L_1$ can recover information about features that were not mentioned~\cite{sumers2021learning, lin2022inferring}. The $L_1$ listener then chooses actions by substituting this posterior belief into Eq.~\ref{eq_listener_policy}. In practice, however, the horizon is not known, and so this approach is not feasible for real-world applications. This model instead serves as a theoretically-optimal baseline, and we next consider three practical ways of handling this uncertainty.

\textbf{Assuming the horizon.} Perhaps the most straightforward approach is to simply assume a speaker horizon. However, prior work has highlighted the risks of assuming a human is actively trying to teach~\cite{milli2020literal}, and suggests that the safest approach is to assume they are not. To test the effects of such mis-specification, in Section~\ref{section_behavioral_data} we define two pragmatic listener models which assume a short ($H=1$) or long ($H=4$) horizon. A pragmatic listener assuming $H=1$ will \emph{constrain} inference: it will assume the utterance reflects only the speaker's preference between actions in the present state, and thus generalize conservatively. In contrast, a pragmatic listener assuming $H\gg1$ expects the utterance to generalize broadly, risking overfitting. 

\textbf{Inferring the horizon.} 
To mitigate the risk of horizon mis-specification, we can instead assume the speaker's horizon is unknown. Given an utterance, the \emph{latent horizon} pragmatic listener jointly infers both their horizon and rewards, then marginalizes out the horizon:
\begin{equation}
    L_1(w \mid s, u) \propto \sum_H S_1(u \mid w, s, H)P(H)P(w)
\label{eq_joint_pragmatic_listener}
\end{equation}
Intuitively, because short-horizon speakers prefer instructions (and long-horizon speakers prefer descriptions, Fig.~\ref{fig:fig2-speaker-horizon-instruction-descriptions}A), the latent-horizon pragmatic listener can use the utterance type to infer the speaker's horizon and determine the appropriate scope of generalization. To demonstrate this, we simulate a pragmatic listener with a uniform prior over $H\in [1, 2, 3, 4, 5, 10]$.  Fig.~\ref{fig:pragmatics} shows example utterances and resulting inference about the speaker's rewards and latent horizon. Crucially, while both example utterances indicate a preference for the spotted red mushroom, the description suggests the speaker is $H>1$ and uniquely identifies the spotted feature as high-value. 

\section{Behavioral experiment}
\label{section_behavioral_data}
To validate our theoretical models, we collected a behavioral dataset. Participants played the role of a mushroom foraging guide and produced utterances for tourists to help them choose good mushrooms. We manipulated the speaker's horizon by varying tourists' itineraries: each tourist was shown visiting a different visible mushroom patch, plus a variable number of unknown future patches (0, 1, or 3, matching horizons 1, 2, and 4 respectively). In the following sections, we analyze the participants' choice of utterances, compare the resulting inference from different listener models, and show how this socially-learned information can accelerate traditional reinforcement learning.\footnote{This study was approved by the Princeton IRB. The full experiment can be viewed at \url{https://pragmatic-bandits.herokuapp.com}.}

\subsection{Experiment setup}
\label{subsec_experiment_setup}

\begin{figure}[]
    \centering
    \includegraphics[width=13.9cm]{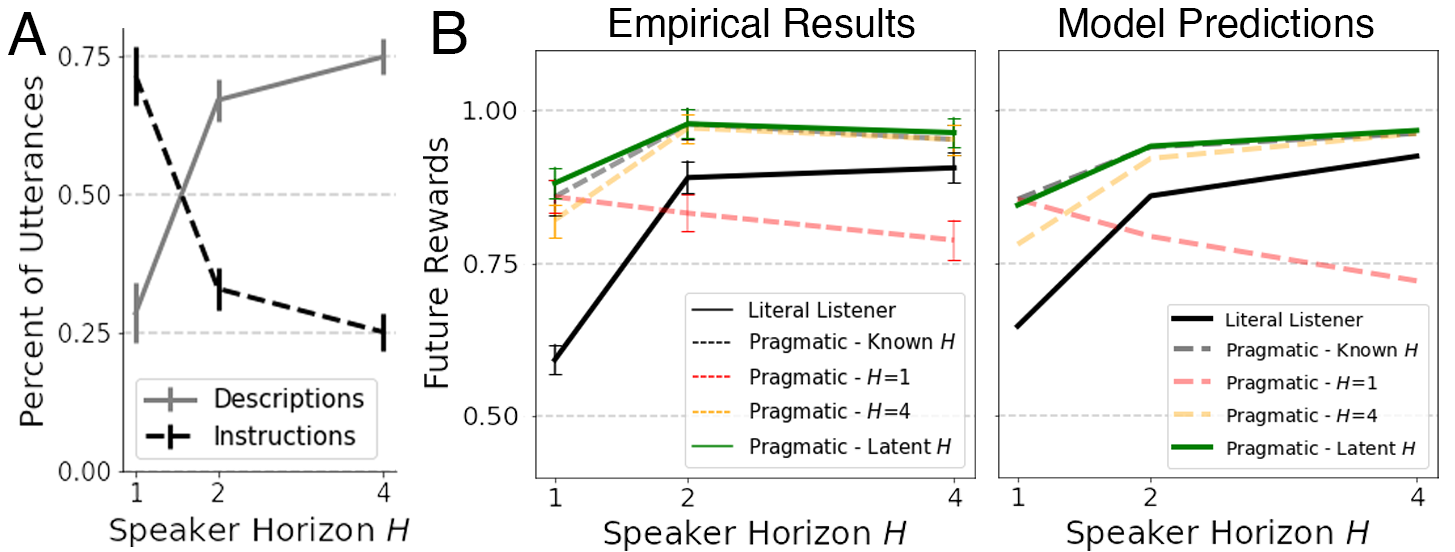}
    \caption{Results from behavioral experiment. \textbf{A}: Participants preferred instructions when $H=1$ and descriptions otherwise. \textbf{B}, left: empirical future rewards (Eq.~\ref{eq_expected_rewards}) from human utterances. As the horizon lengthens, humans choose more generalizable utterances; our latent-horizon listener successfully recovers their reward function. Right: simulations using our speaker model with $\beta_{S_1}=3$. All error bars show 95\% CI. }
    \label{fig:behavioral_data}
\end{figure}

We recruited 119 participants on the Prolific experimental platform (\url{prolific.co}). Participants were trained and tested on the game dynamics then advised a total of 28 tourists. They were told to consider the tourists' itinerary and choose ``the most helpful utterance'' from drop-down menus allowing them to specify an instruction or description. Because the space of possible descriptions (30) is larger than possible instructions (9), choosing a description required more effort than choosing an instruction. To equalize this, we reduced the set of descriptions by removing neutral features (``red'' and ``solid'') and the 0 value, yielding $4\times4=16$ possible descriptive utterances. For the remainder of the paper, all pragmatic listeners assume the speaker chooses from this reduced utterance set. Participants received no feedback on how their utterances affected tourists' behaviors, ensuring they chose utterances according to their own sense of how to help. After screening out participants who failed comprehension or attention checks, we were left with 99 participants who produced a total of 2772 utterances. For more details on the data collection, see Appendix~\ref{subsec_appdx_experiment_details}.

We next use this set of utterances to explore value alignment (whether our models can infer the speakers' reward function) and then integrated social and reinforcement learning (whether this socially-acquired information can reduce regret in traditional RL).

\subsection{Inferring rewards: value alignment from language}
\label{subsec_behavioral_reward_inference_results}
Overall, our results validated our theoretical models. First, participants were sensitive to the horizon manipulation: there was a statistically-significant shift in their utterance choices between horizons (for $H=1$ to $2$, $\chi^2(26, 1856) = 336.1, p<.001$; for $H=2$ to $4$, $\chi^2(26, 1833) = 40.3, p=.04$). Almost all participants (96 out of 99) used a mix of instructions and descriptions, favoring instructions at $H=1$ and descriptions at $H>1$ (Fig.~\ref{fig:behavioral_data}A). This led to lower literal future rewards when $H=1$, and higher future rewards at longer horizons (Fig~\ref{fig:behavioral_data}B, ``Empirical Results''). To calibrate our pragmatic listeners, we tested $\beta_{S_1} \in [1, 10]$ and found that $\beta_{S_1}=3$ optimized Known $H$ and Latent $H$ listeners (see Appendix~\ref{subsec_appdx_choosing_beta} for details). Simulating utterances produced by our speaker model and resulting pragmatic inference shows a close match to theoretical predictions (Fig.~\ref{fig:behavioral_data}B, ``Model Predictions'', see Appendix~\ref{subsec_appdx_simulating_human_behavioral_data} for details). 

Our pragmatic listeners offered statistically-significant improvements over the literal listener (Table~\ref{tab:results_table}). These gains were particularly large when the speaker has a short horizon, matching our simulations. Somewhat surprisingly, our Latent $H$ model ($M=0.94$, $SD=0.39$) outperformed all other models, including the Known $H$ model ($M=.93$, $SD=.40$), by a significant albeit small margin (mean difference 0.01, paired-samples t-test $t(2771)=4.18, p<.001$; see Appendix~\ref{sec_appdx_statistical_testing} for other pairwise tests). This is particularly notable because it underscores the inevitability of misspecification~\cite{milli2020literal}: even when we experimentally controlled the horizon, participants did not perfectly follow our theory, leading our Latent $H$ model to outperform. We also confirmed the risks of \emph{assuming} a horizon: the two fixed-horizon listeners ($H=1$ and $H=4$) underperformed the Latent $H$ listener. 

Despite these successes, we found a notable discrepancy between theoretical predictions and empirical results, suggesting a possible future refinement. As described in~\cite{sumers2021extending}, utility-maximizing speakers agnostic to the truth of an utterance may send messages with exaggerated values (e.g. preferring the false utterance $\langle$Spotted, +2$\rangle$ to the true utterance $\langle$Spotted, +1$\rangle$). In our experiment, however, participants regularly chose the lower-reward, true utterance (Appendix~\ref{sec_appdx_behavioral_exp}). Breaking out results by utterance type indicates that pragmatic gains come primarily on instructions (Appendix~\ref{sec_appdx_pragmatic_breakdown}). This suggests that our reward-design objective is in fact too weak: participants' tendency to use true descriptions licenses stronger inference, which we return to in the discussion.

\subsection{Reducing regret: integrating social and reinforcement learning}
\label{subsec_behavioral_rl_integration_results}
\begin{figure}
\begin{minipage}{\textwidth}
\begin{minipage}[b]{0.45\textwidth}
\centering
    \includegraphics[width=6cm]{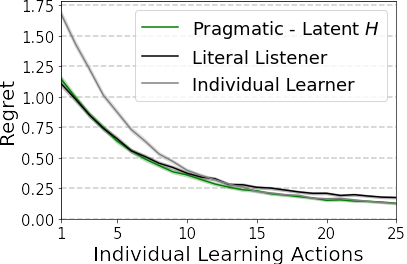}
\captionof{figure}{Regret with and without social information from our behavioral experiment.}
\label{fig:individual_learning}
\end{minipage}
\hfill
\begin{minipage}[b]{0.54\textwidth}
\centering
\begin{tabular}{lcc}
\toprule
\multirow{2}{*}{\textbf{Listener}} & Future rewards        & Regret after     \\
                                   & from utterance &  25 actions  \\ \hline
Individual                       & -                                            & 12.14           \\
Literal                          & .79                                                   & 10.23           \\ 
Prag - Known $H$                       & .93                                  & \textbf{9.42}*   \\
Prag - $H = 1$                         & .83                                          & 9.76            \\
Prag - $H = 4$                         & .91                                          & 9.67            \\
Prag - Latent $H$                      & \textbf{.94}***                                & 9.55         \\  
\bottomrule
\end{tabular}
  \captionof{table}{Results from human utterances. Higher rewards (Section~\ref{subsec_behavioral_reward_inference_results}, Eq.~\ref{eq_expected_rewards}) and lower regret (Section~\ref{subsec_behavioral_rl_integration_results}) are better. Bolded results are significantly better. }
  \label{tab:results_table}
\end{minipage}
\end{minipage}
\end{figure}

We now explore how this form of social learning could augment traditional RL approaches. In particular, we take the inferences from our listeners in the previous section and use them as a prior for reinforcement learning (or in the case of literal instructions, we integrate them into the learner's policy; see Appendix~\ref{sec_appdx_thompson_sampling} for details on the experimental setup). We then use Thompson sampling~\cite{thompson1933thompson, russo2014thompson, russo2018thompson} to learn the reward function and compare regret over the course of learning for five independent learning trials on each context-utterance pair. We report results for our listener agents, as well as an agent with no social information (the ``Individual'' agent; Fig.~\ref{fig:individual_learning}, Table~\ref{tab:results_table}). 

We find that social information can substantially reduce regret. Intriguingly, when comparing regret, our Known $H$ pragmatic listener now achieves the best results; this is likely due to the greater uncertainty inherent in the Latent $H$ inference. To confirm these differences are statistically significant, we use a linear regression to predict regret with a fixed effect of listener and random effects for each of the 2772 utterance-context pairs, again comparing the Latent $H$ listener to all other listeners. This confirms that the Known $H$ listener achieves lower regret than the Latent $H$ listener ($\beta = -.12$, $t(80400)=-2.14, p=.03$), and all other models suffer higher regret (see Appendix~\ref{sec_appdx_statistical_testing}).

\section{Discussion}
\label{section_discussion}
We formalized the challenge of using language to teach an agent to act on our behalf~\cite{HadfieldMenell2016cooperative, fisac2020pragmatic, milli2017should, amin2017repeated, jeon2020reward} in a bandit setting, introducing the notion of a \emph{horizon} to reflect the agent's degree of autonomy. We find \emph{instructions} are optimal for low-autonomy settings, while \emph{descriptions} are optimal for high autonomy. This distinction allows a pragmatic listener to jointly infer the speaker's horizon and reward function, reducing misspecification risk~\cite{milli2020literal}. Our behavioral experiment supports our theory, demonstrating the benefit of learning from language for value alignment~\cite{Gabriel2020alignment, russell2019humancompatible, amodei2016concrete} and traditional RL~\cite{sutton2018reinforcement}.

We note several limitations and future directions. First, human participants preferred \emph{truthful} descriptions; this licenses stronger pragmatics~\cite{shah2019biases} such as combining our model with epistemic objectives~\cite{frank2012pragmatic, goodman_2016}. Second, we assumed language groundings were known, but future work could incorporate uncertainty~\cite{degen2020redundancy, hawkins2022partners} or learn them~\cite{mooney2008learning, sumers2021learning, lin2022inferring}. Third, we studied a simple contextual bandit setting;  sequential decision settings will require models of planning, and may consider additional language such as instructions at different levels of abstraction or \emph{transition} descriptions~\cite{Rafferty2011faster}. Lastly, we modeled a simple interaction consisting of a single utterance. More naturalistic interactions would allow multiple utterances, bidirectional dialogue, or interleaved action and communication~\cite{HadfieldMenell2016cooperative, fisac2020pragmatic}.

Our work can inform agent design by clarifying how and when different forms of linguistic input can be useful.
Instruction following may be optimal when autonomy is unnecessary or preferences are non-Markovian~\cite{abel2021expressivity}. 
However, if autonomy is desired~\cite{Gabriel2020alignment, russell2019humancompatible, amodei2016concrete, milli2017should, milli2020literal, jeon2020reward, HadfieldMenell2016cooperative, fisac2020pragmatic}, agents should be equipped to understand descriptions of the world or our preferences~\cite{narasimhan2018grounding, zhong2019rtfm, hanjie2021grounding, sumers2021learning, lin2022inferring}. 
Finally, pragmatic agents can \emph{infer} whether preferences are local or general. This suggests that learning from a wide range of language is a promising approach for both value alignment and RL more broadly.

\begin{ack}
We thank Rachit Dubey, Karthik Narasimhan, and Carlos Correa for helpful discussions. TRS is supported by the NDSEG Fellowship Program and RDH is supported by the NSF (grant \#1911835).
This work was additionally supported by a John Templeton Foundation grant to TLG (\#61454) and a grant from the Hirji Wigglesworth Family Foundation to DHM.
\end{ack}

\bibliography{references}

\pagebreak
\appendix

\renewcommand{\thefigure}{S\arabic{figure}}
\renewcommand{\thetable}{S\arabic{table}}
\setcounter{table}{0}
\setcounter{figure}{0}

\section{Speaker simulations and pragmatic inference}
\label{sec_appdx_pragmatic_simulations}

\subsection{Instructions vs Descriptions}
In the main text, we used a fixed number of available actions (a context $S$ with $|S|=3$ objects).
Here, we further explore the effect of horizon on the choice of instructions vs. descriptions under different numbers of available actions. 
As in the main text, we assume that the speaker uses a near-optimal softmax temperature for clarity by setting $\beta_{S_1} = 10$. 

\begin{figure}[h]
    \centering
    \includegraphics[width=5cm]{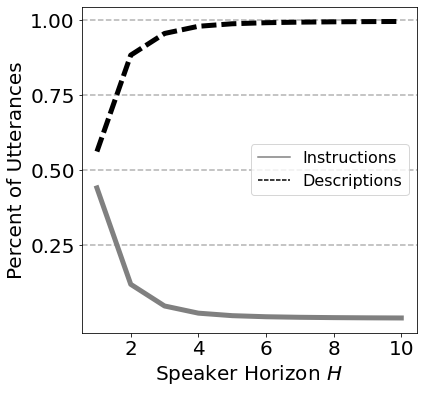}
    \caption{As noted in Section~\ref{subsec_instructions_vs_descriptions}, speakers exhibit a strong preference for descriptions as their horizon lengthens.}
    \label{fig:instructions-vs-descriptions-horizon}
\end{figure}
\FloatBarrier

\begin{figure}[h]
    \centering
    \includegraphics[width=13.9cm]{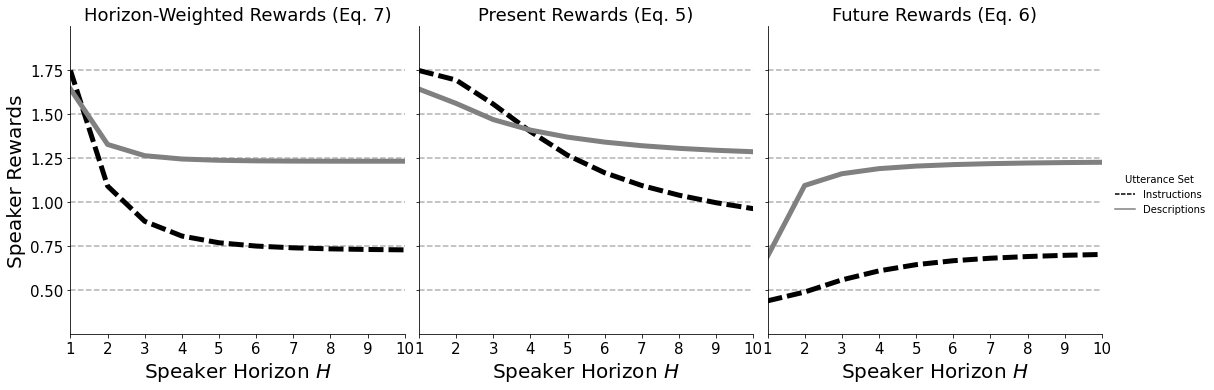}
    \caption{Breaking out speaker rewards (Fig.~\ref{fig:fig2-speaker-horizon-instruction-descriptions}B) by reward type and utterance. ``Horizon-Weighted Rewards'' (left) is the same as Fig.~\ref{fig:fig2-speaker-horizon-instruction-descriptions}B. Instructions afford high ``Present Rewards'' (center) but generalize poorly (low ``Future Rewards'', right). As a result, rational speakers with access to instructions only remain biased towards the present context even as their horizon lengthens. This can be seen by comparing ``Present'' and ``Future'' rewards at long horizons (e.g. $H=10$). Description-only speakers exhibit little bias towards the present context (``Present'' and ``Future'' rewards are nearly equal), while instruction-only speakers remain biased towards the present context (``Present'' > ``Future'' rewards).}
    \label{fig:present-vs-future-rewards}
\end{figure}
\FloatBarrier

We find that instructions become more useful as the number of available actions increases. They can always uniquely select the best action in a given state (even when all nine possible objects are present), whereas it is not always possible to use a description to identify the best action.

\begin{figure}[h]
    \centering
    \includegraphics[width=11cm]{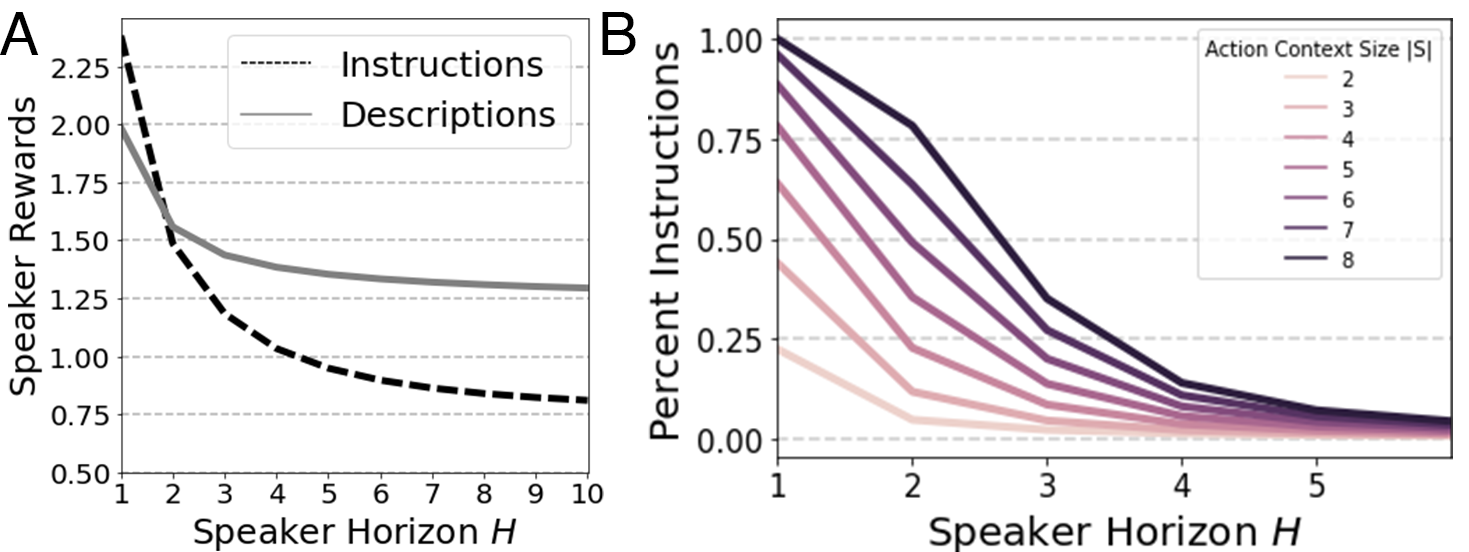}
    \caption{\textbf{A}: Same as Fig.~\ref{fig:fig2-speaker-horizon-instruction-descriptions}B, but with a state-size $|S|$ of 5 instead of 3. At short horizons, the relative utility of instructions increases with state size (e.g. as the action space grows, instructions are more useful). \textbf{B}: Speaker's probability of using an instruction as a function of number of available actions $|S|$ and horizon $H$ (note that Fig.~\ref{fig:instructions-vs-descriptions-horizon} shows the curve for $|S|=3$). As the number of actions increases, speakers prefer instructions.}
\end{figure}
\FloatBarrier

Next, building on Fig.~\ref{fig:pragmatics} in the main text, we include full posteriors over reward functions and additional utterances. 

\begin{figure}[h]
    \centering
    \includegraphics[width=13.9cm]{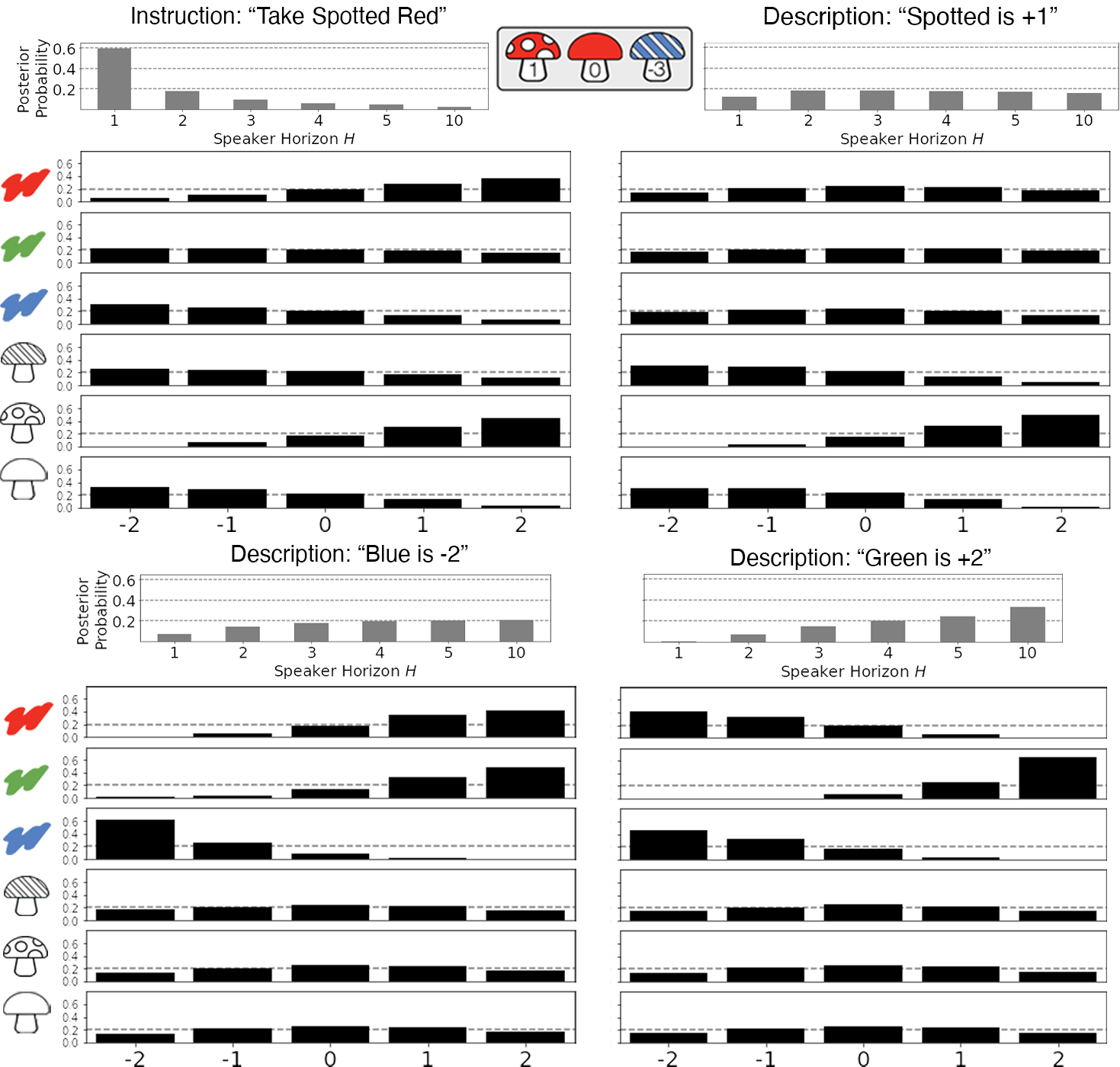}
    \caption{Same as Fig.~\ref{fig:pragmatics}, but with full utterance posteriors over possible reward functions and additional utterances. For utterance posteriors, the gray dashed line indicates the prior (e.g. uniform over all possible values). Note that descriptions suggest that unmentioned features are lower-magnitude (e.g. for the bottom-right ``Green is +2'' utterance, the listener infers that all textures---Striped, Spotted, and Solid---are unlikely to be -2 or +2). Finally, note that with all descriptive utterances, the listener assigns non-negligible probability mass to values \emph{other} than the specified one (e.g. in the top-right, the listener infers a substantial probability that the Spotted feature is actually +2). This suggests that integrating a ``truthfulness bias'' could improve our models (see Section~\ref{subsec_behavioral_reward_inference_results} for a discussion).}
    \label{fig:appendix_pragmatic_posteriors}
\end{figure}
\FloatBarrier

\section{Behavioral Experiment}
\label{sec_appdx_behavioral_exp}
The full experiment can be viewed at \url{pragmatic-bandits.herokuapp.com}. Note that the app is running on free dynos so you may need to wait 5-10 seconds for it to load.

\subsection{Experiment Details}
\label{subsec_appdx_experiment_details}
\textbf{Participant compensation.} Participants earned an average hourly wage of \$12.10 and the total amount spent on participant compensation was \$444. The mean time spent in the experiment was 18.5 minutes. 

\textbf{IRB approvals.} This study was approved by the Princeton University IRB. All participants gave informed consent; the consent form can be seen at the experiment URL above. As described in the Checklist, no significant participant risks were anticipated.

\textbf{Anonymized data.} Anonymized data, including participant responses and free-form exit survey responses, is available in the supplementary zip file and will be released along with the code for this paper. Note that the worker IDs provided have been hashed to prevent re-identification of participants on the platform. 

\textbf{Trials.}
We split our 84 states into 3 sets of 28; each participant saw one of these sets. Each participant additionally saw 8 ``attention check'' trials (constant across all participants). These ``attention checks'' forced the participant to use a description with a pre-selected feature (4 ``Spotted'' and 4 ``Striped''). The participant then chose a value from $[-1, +1]$. The trials were selected to ensure that the true value would lead the learner to choose a good mushroom. Participants who failed to select the true value on at least 7/8 trials (e.g. >75\% of the time) were still paid the full bonus, but their data was excluded from the analysis. We further exclude the attention check trials from the analysis.

\begin{figure}[h]
    \centering
    \includegraphics[width=10cm]{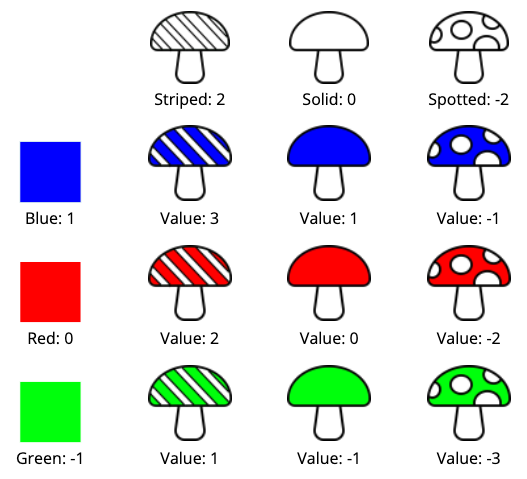}
    \caption{Throughout the experiment, participants could trigger a pop-up giving them the value of all features (and all actions). Note that features were randomized to avoid saliency biases, and so this set of feature values does not match Fig.~\ref{fig:linear_bandits_setup}.}
    \label{fig:exp-mushroom-cheat-sheet}
\end{figure}

\textbf{Feature Randomization.}
To avoid saliency biases (e.g. color may be more salient than texture), mushroom feature values were randomized across participants. Fig.~\ref{fig:exp-mushroom-cheat-sheet} shows one example of an alternative featurization scheme. Note that all responses were converted back the the ``canonical'' feature map shown in Fig.~\ref{fig:linear_bandits_setup} for analysis.

\FloatBarrier

\textbf{Instructions} We include several screenshots of key instruction pages, but recommend viewing the full experiment at \url{pragmatic-bandits.herokuapp.com} for details.

\begin{figure}[h]
    \centering
    \includegraphics[width=8cm]{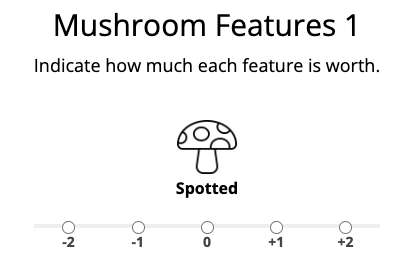}
    \caption{One example quiz question. To ensure comprehension of the contextual bandit setup, participants were tested on their knowledge of all features. If they failed the quiz, the experiment terminated early and they earned \$2; if they passed, they completed the full experiment and earned \$4.}
    \label{fig:exp-mushroom-feature-quiz}
\end{figure}
\FloatBarrier

\begin{figure}[h]
    \centering
    \includegraphics[width=13.9cm]{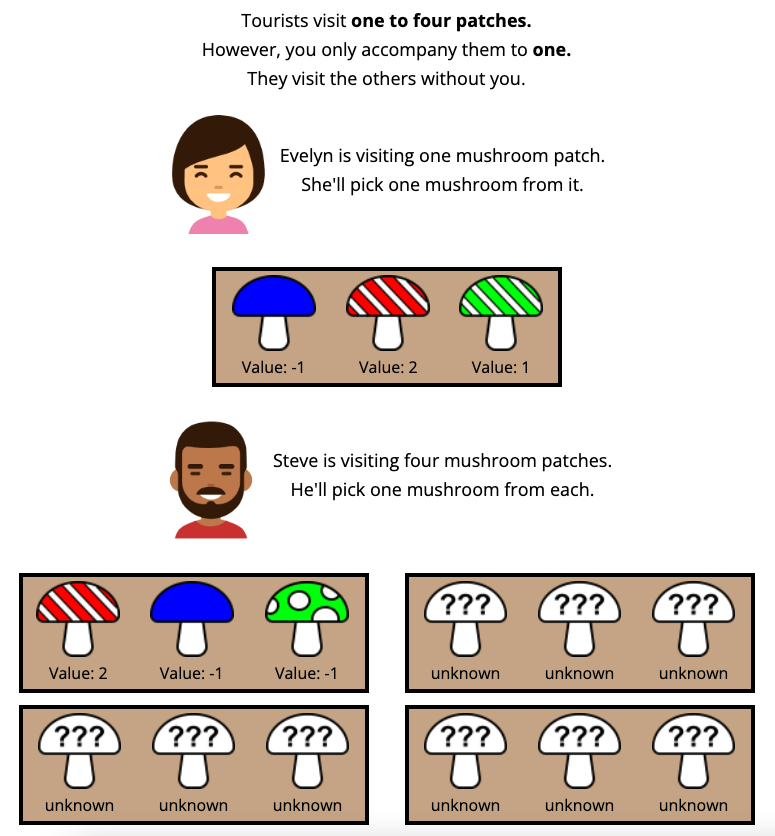}
    \caption{Experiment instructions: introducing the notion of the \emph{horizon}. Participants were told they could only accompany the tourist to one patch, but that depending on their itinerary, tourists could go on to visit other (unknown) patches afterwards.}
    \label{fig:exp-instructions-horizon}
\end{figure}
\FloatBarrier

\begin{figure}[h]
    \centering
    \includegraphics[width=13.9cm]{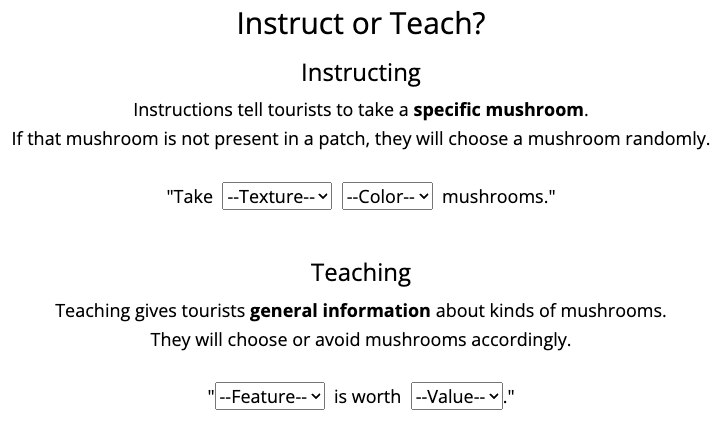}
    \caption{Experiment instructions: introducing the notion of instructions and descriptions. Participants were told to consider the tourist's itinerary (Fig.~\ref{fig:exp-instructions-horizon}) and help the tourist pick good mushrooms throughout their visit.}
    \label{fig:exp-instructions-utterances}
\end{figure}

\begin{figure}[h]
    \centering
    \includegraphics[width=13.9cm]{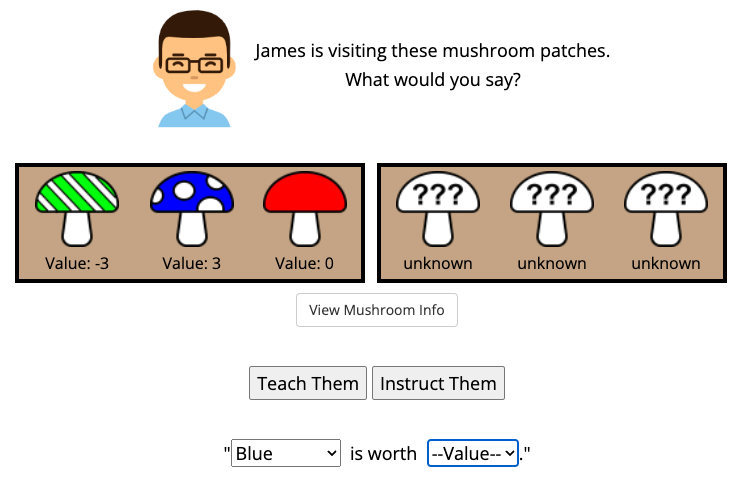}
    \caption{One example trial from the experiment. Clicking the ``Teach Them'' button revealed drop-down menus to select a ``Description'' utterance, while clicking ``Instruct Them'' yielded menus to select an ``Instruction.''}
    \label{fig:exp-example-trial}
\end{figure}
\FloatBarrier

\subsection{Participant Utterance Choices}
\label{subsec_appdx_utterance_choices}

The supplementary materials contain all participant responses (see the Experiment Analysis Jupyter notebook for analysis code). Here, we summarize some of the key patterns in the data.

\begin{figure}[h]
    \centering
    \includegraphics[width=8cm]{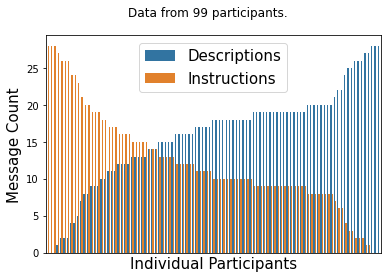}
    \caption{Breakdown of utterance types (instruction vs descriptions) for all participants. Most participants used a mix of both: 3 used only instructions, 3 used only descriptions, and 93 used at least one message of each type.}
\end{figure}
\FloatBarrier

\begin{figure}[h]
    \centering
    \includegraphics[width=12cm]{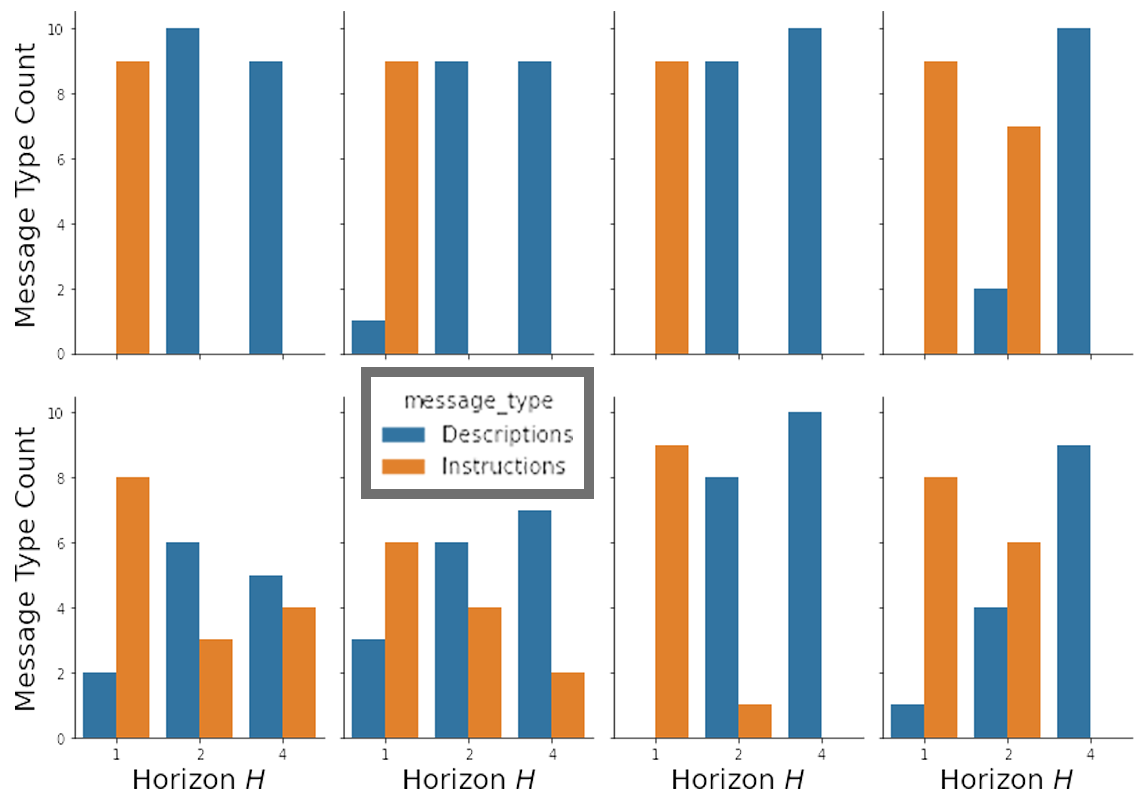}
    \caption{Instruction / description breakdown by horizon for 8 random participants. While individual preferences varied substantially, virtually all participants displayed an increasing preference for descriptions as the horizon increased.}
\end{figure}
\FloatBarrier

\begin{figure}[h]
    \centering
    \includegraphics[width=13.9cm]{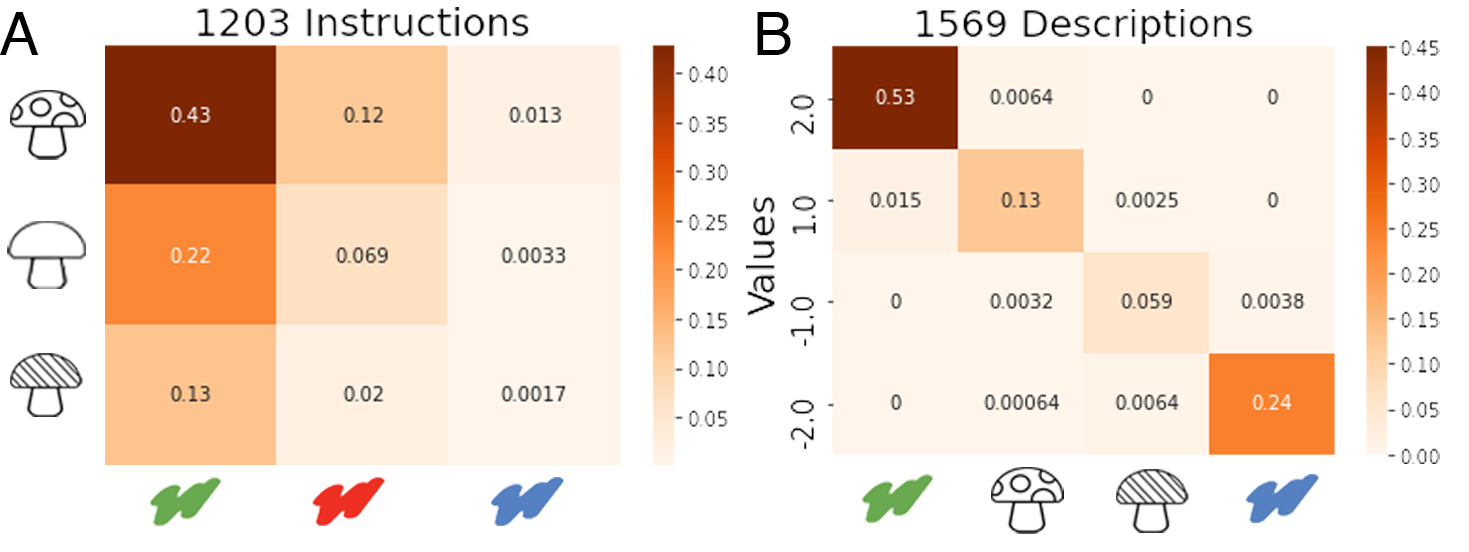}
    \caption{Within-type distribution of utterances chosen by participants. \textbf{A}: When giving instructions participants (unsurprisingly) almost always chose positive-reward actions, e.g. ``Take the spotted green mushroom,'' in the top-left quadrant. \textbf{B}: When giving descriptions, participants almost always chose \emph{true} utterances (e.g. ``Spotted is +1''), even though our reward-maximizing model predicts exaggeration (e.g. ``Spotted is +2''). See Section~\ref{subsec_behavioral_reward_inference_results} for discussion.}
    \label{fig:utterance-heatmaps}
\end{figure}
\FloatBarrier

\subsection{Choosing $\beta_{S_1}$}
\label{subsec_appdx_choosing_beta}

To choose a $\beta_{S_1}$ for our behavioral experiment, we used a grid search over integers $\beta_{S_1} \in [1, 10]$ and evaluated our primary models (Pragmatic - Known $H$ and Pragmatic - Latent $H$). We chose $\beta_{S_1}=3$, which optimized future reward from the human data for both speakers. Note that while ``Pragmatic - Known $H$'' was numerically optimal at $\beta_{S_1} = 2\text{ (expected reward}=0.9308, SD= 0.39)$, there was not find a significant difference between this and $\beta_{S_1} = 3\text{ (expected reward}=0.9295, SD=0.40)$; paired-samples $t(2771) = -1.70, p=0.09$.

\begin{figure}[h]
    \centering
    \includegraphics[width=13.9cm]{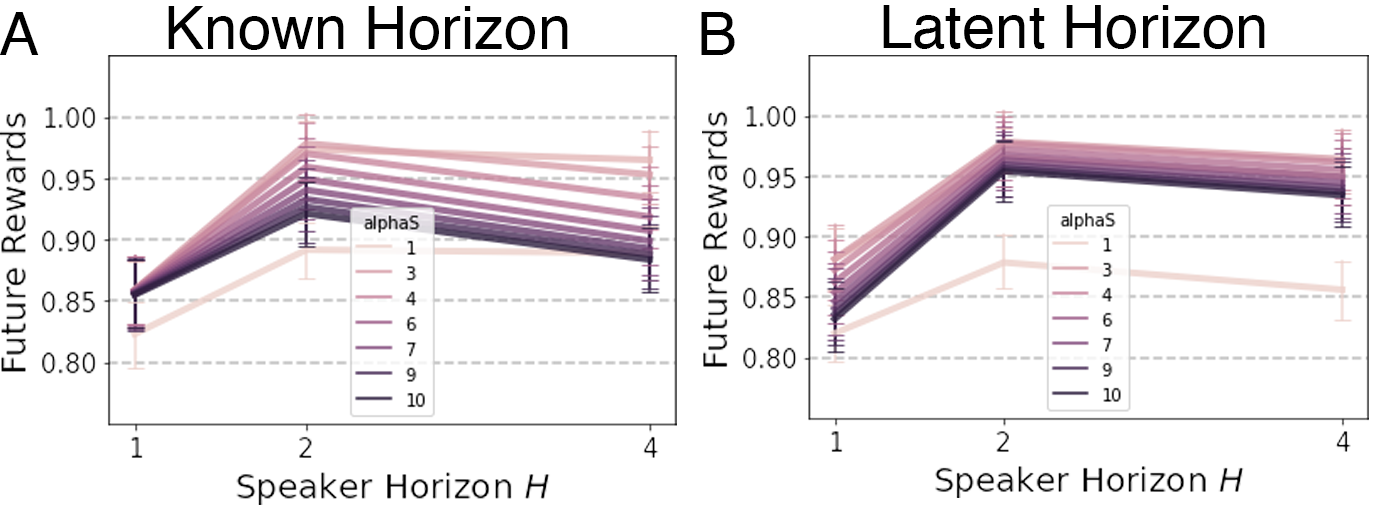}
    \caption{Performance of different pragmatic listener models as a function of horizon $H$ and speaker-optimality $\beta_{S_1}$. Qualitatively, the Latent-$H$ model (right) was less sensitive than the Known-$H$ model (left).}
\end{figure}
\FloatBarrier




\begin{table}
\centering
\begin{tabular}{c|cc}
$\beta_{S_1}$ & Known $H$ & Latent $H$  \\ \hline
1 & 0.87 & 0.85 \\
2 & 0.93 & 0.93 \\
\textbf{3} & \textbf{0.93} & \textbf{0.94} \\
4 & 0.92 & 0.94 \\
5 & 0.91 & 0.93 \\
6 & 0.90 & 0.92 \\
7 & 0.90 & 0.92 \\
8 & 0.89 & 0.91 \\
9 & 0.89 & 0.91 \\
10 & 0.89 & 0.91 \\
\end{tabular}
\caption{Mean ``Future Rewards'' for our two primary models of interest as a function of the $\beta_{S_1}$ parameter. Note that while ``Pragmatic - Known $H$ was slightly higher at $\beta_{S_1} = 2$ there was not find a significant difference between this value and $\beta_{S_1} = 3$, hence we use the latter to be consistent across both models.}
\centering
\end{table}
\FloatBarrier

\subsection{Simulating model behavior}
\label{subsec_appdx_simulating_human_behavioral_data}
To compare the empirical pattern of utterances observed from humans in our experiment against the predictions of our theoretical speaker model, we use simulations to generate a distribution over utterances and directly compare the results (Fig.~\ref{fig:behavioral_data}, ``Model Predictions''). We set $\beta_{S_1}=3$ as described above. The utterance set is composed of the 9 instructions and 16 descriptions defined in Section~\ref{subsec_experiment_setup}, for a total of 25 possible utterances.
 
First, for each $H\in [1, 2, 4]$ and each of the 84 states $s\in S$, we run the speaker model to produce a distribution over the 25 possible utterances (Eq.~\ref{eq_speaker_combined_utility}). We then calculate the literal future rewards resulting from each utterance (Eq.~\ref{eq_literal_listener_instruction} for instructions and Eqs.\ref{eq_literal_listener_belief_update},~\ref{eq_listener_policy} for descriptions). 
We then calculate the \emph{expected future reward} obtained by the literal listener by weighting the rewards for each utterance by the probability of the speaker producing that utterance, and averaging over all 84 start states. 

Similarly, to evaluate the pragmatic listener, we perform pragmatic inference over each utterance (Eq.~\ref{eq_fixed_horizon_pragmatics}) to recover the speaker's reward function, evaluate the future rewards (Eq.~\ref{eq_expected_rewards}) from the resulting beliefs, and again weight by the speaker's distribution over utterances.  

\section{Statistical testing}
\label{sec_appdx_statistical_testing}
See the R notebook for statistical testing code. 
\subsection{Paired T-Tests (\S~\ref{subsec_behavioral_reward_inference_results})}

\begin{table}[h!]
\begin{tabular}{l|l|cl|l|l|c}
\multicolumn{1}{c|}{Comparison} & \multicolumn{1}{c|}{Mean Difference} & \multicolumn{2}{c|}{95\% CI} & \multicolumn{1}{c|}{t} & \multicolumn{1}{c|}{df} & p-val                            \\ \hline
vs. Pragmatic (Known $H$)                & 0.011                                & 0.006         & 0.017        & 4.1798                 & 2771                    & <.001                        \\
vs. Pragmatic ($H=4$)                    & 0.026                                & 0.022         & 0.031        & 10.81                  & 2771                    & <.001 \\
vs. Pragmatic ($H=1$ )                 & 0.11                                & 0.10         & 0.13        & 20.545                 & 2771                    & <.001 \\
vs. Literal                         & 0.15                                & 0.13         & 0.16        & 23.364                 & 2771                    & <.001
\end{tabular}
\caption{Pairwise $t$-tests comparing the ``Future Rewards'' obtained by the Latent-$H$ listener to other models for the 2772 utterances from our behavioral experiment. These results indicate that the Latent $H$ model outperforms all other models.}
\end{table}

\FloatBarrier
     
\begin{table}[h!]
\begin{tabular}{l|c|ll|c|c|c}
Comparison       & Mean Difference & \multicolumn{2}{c|}{95\% CI} & t       & df   & p-val             \\ \hline
vs. Pragmatic (Known $H$) & -0.13          & -0.15        & -0.12       & -19.854 & 2771 & \textless .001 \\
vs. Pragmatic ($H=4$)     & -0.12          & -0.13        & -0.11      & -23.324 & 2771 & \textless .001 \\
vs. Pragmatic ($H=1$)     & -0.03          & -0.05        & -0.01       & -3.202  & 2771 & <.01        
\end{tabular}
\caption{Pairwise $t$-tests comparing the ``Future Rewards'' obtained by the Literal listener to the remaining other models for the 2772 utterances from our behavioral experiment. These results indicate that all pragmatic models outperform the Literal model.}
\end{table}
\FloatBarrier

\subsection{Mixed-effects regression model (\S~\ref{subsec_behavioral_rl_integration_results})}
The following analysis tests for a significant difference in regret when using the model's social-learning posterior as a prior for individual learning (see Section~\ref{sec_appdx_thompson_sampling} for details). Note that lower regret is better, so negative coefficients indicate better performance.

We dummy-coded our different models as a categorical variable with the Latent $H$ listener as the reference level. We included random intercepts for each unique utterance from our experiment  (e.g. for each of the 2772 utterances chosen by participants) to account for some utterances being systematically easier or harder than others. 
The resulting coefficients indicate that the Latent $H$ listener outperformed all models except for the Known $H$ model, which achieved slightly lower regret.

\begin{table}[ht]
\centering
\begin{tabular}{rllrrrrrr}
  \hline
 & Effect & Term & Estimate & Std Error & Statistic & df & $p$ value \\ 
  \hline
1 & fixed &   (Intercept) & 9.55 & 0.05 & 195.66 & 14282.57 & $<0.001$ \\ 
  2 & Fixed   & Individual & 2.60 & 0.06 & 45.41 & 80383.00 & $<0.001$ \\ 
  3 & Fixed   & Literal & 0.68 & 0.06 & 11.93 & 80383.00 & $<0.001$ \\ 
  4 & Fixed   & Prag (Known  $H$) & -0.12 & 0.06 & -2.14 & 80383.00 & 0.03 \\ 
  5 & Fixed   & Prag ($H=1$) & 0.22 & 0.06 & 3.77 & 80383.00 & $<0.001$ \\ 
  6 & Fixed   & Prag ($H=4$) & 0.13 & 0.06 & 2.24 & 80383.00 & 0.03 \\ 
  7 & Random Effect & sd\_\_(Intercept) & 1.44 &  &  &  &  \\ 
  8 & Residual & sd\_\_Observation & 4.76 &  &  &  &  \\ 
   \hline
\end{tabular}
\end{table}
\FloatBarrier

\section{When does pragmatic reasoning help?}
\label{sec_appdx_pragmatic_breakdown}
In this section, we examine the utterances produced in the human experiment (Section~\ref{section_behavioral_data} and Appendix~\ref{sec_appdx_behavioral_exp}) to explore when, exactly, pragmatic reasoning is most useful.
We analyze the performance of the Latent $H$ pragmatic model in comparison to the Literal listener. 
Concretely, we take the 2772 utterances produced in our behavioral experiment and evaluate the ``future rewards'' (Eq.~\ref{eq_expected_rewards}, the expected rewards over all possible states, ) resulting from a literal interpretation of the utterance against those resulting from a pragmatic interpretation.

\begin{table}[h]
\centering
\begin{tabular}{l|cc}
Utterance Type & Count & Mean Pragmatic Gain \\ \hline
Instruction & 1203 & .42 $\pm$ .23 \\
Description & 1569 & -.07 $\pm$ .22
\end{tabular}
\caption{Average pragmatic gain for different utterance types (+/- standard deviations). Pragmatics on instructions helps substantially by converting from partial policies to rewards, but pragmatics on descriptions marginally \emph{reduces} the average reward obtained.}
\label{tab:appdx_pragmatics_by_utterance_type}
\end{table}
\FloatBarrier

We find that under these conditions, pragmatic inference primarily helps with \emph{instructions}, rather than descriptions (Table~\ref{tab:appdx_pragmatics_by_utterance_type}): converting a partial policy into inference over the reward function allows much stronger generalization. Across the 1203 instruction utterances in our experiment, the pragmatic listener achieved a large and statistically-significant gain ($M=.423,SD=.232$), $t(1202)=63.34, p<.001$. In contrast, on the 1569 descriptive utterances, the pragmatic listener suffered a small but statistically-significant loss ($M=-.067,SD=.215$), $t(1568)=-12.29, p<.001$. 

\begin{figure}[h]
    \centering
    \includegraphics[width=7cm]{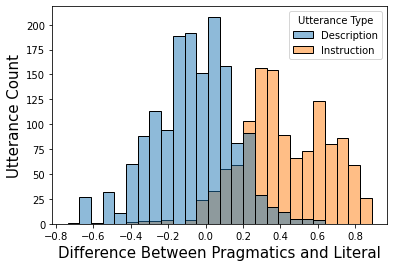}
    \caption{Distribution of pragmatic gain (Pragmatic listener with Latent $H$ vs.  Literal listener) for the 2772 utterances in our behavioral experiment. Pragmatic inference substantially improves rewards for instructions, but marginally reduces rewards for descriptions on expectation.}
    \label{fig:appdx_literal-pragmatic-instruction-description-comparison}
\end{figure}
\FloatBarrier


Analysis of utterance posteriors (Fig.~\ref{fig:appendix_pragmatic_posteriors}) shows one notable disconnect with empirical human behavior regarding descriptive utterances. The pragmatic listener \emph{does not} preserve the literal truth conditions of descriptive utterances: for the three descriptions shown in Fig.~\ref{fig:appendix_pragmatic_posteriors}, the listener places substantial probability mass on values \emph{other} than the specified one (e.g. believing that ``Spotted is +1'' suggests ``Spotted is +2'' is plausible). Yet in our experiment, participants almost always choose true utterances (see Fig.~\ref{fig:utterance-heatmaps}B). This suggests future work integrating truthfulness and reward objectives, effectively combining our current objective with classic Gricean notions.

\section{Social vs. individual reinforcement learning (\S~\ref{subsec_behavioral_rl_integration_results})}
\label{sec_appdx_thompson_sampling}
To study the potential benefits of integrating social and reinforcement learning, we integrated the reward information learned from our behavioral experiment into a classic Thompson sampling individual learner in \S~\ref{subsec_behavioral_rl_integration_results}. Here, we provide details on this integration.
Code for these simulations can be found in the Supplemental Materials.

\subsection{Individual learning: Thompson sampling in contextual bandits}
\label{sec_appdx_individual_learning_thompson_sampling}

We first define a simple individual learner in our linear contextual bandit setting using Thompson sampling~\cite{thompson1933thompson, russo2014thompson, russo2018thompson}. The agent begins with a prior distribution over possible reward functions. At each timestep, they (1) observe a new state $s_t$ consisting of three possible actions; (2) sample a reward weight vector $w_t$ from their belief distribution over reward weights, and (3) act optimally according to that reward vector. They observe the reward of that action, and use this observation to update their beliefs for the next timestep.

We implement this algorithm using a Gaussian prior and likelihood function, assuming observation noise from a unit Gaussian. Thus, after taking action $a$, the agent receives rewards according to:
\begin{equation}
    R(a) \sim \mathcal{N}(\phi(a)^{\top} w, 1)
    \label{bayesian_lr_equation}
\end{equation}

We use a wide multivariate Gaussian prior: $w^0 \sim \mathcal{N}(0, \Sigma_0)$ where $\Sigma_0 = 5I$. After each action, we perform conjugate Bayesian updates to obtain a posterior (i.e. use Bayesian linear regression), which we use for for the next timestep.

We perform rejection sampling to ensure the sampled belief is compatible with the (discrete, bounded) reward function. We first sample a (continuous) weight vector from our multivariate Gaussian beliefs, then round the weights to integer values and reject the sample if any of the resulting values fall outside the range of possible reward values, $[-2, 2]$. 

We note that these simulation parameters are arbitrary.  Our aim is to demonstrate the general utility of social information to reduce regret \emph{even when individual learning is entirely possible}. We thus defined a relatively straightforward, low-noise individual learning setting. However, we could easily make individual learning arbitrarily more difficult (e.g. by increasing the observation noise), which would in turn increase the relative value of social information.

\subsection{Integrating pragmatic inference: Importance sampling}

In order to integrate social information about the reward function, we incorporate an additional importance sampling step.
Given a particular pragmatic model and a utterance-context-horizon tuple, we first use the pragmatic model to generate a \emph{social} posterior over reward functions (Eq.~\ref{eq_fixed_horizon_pragmatics}~or~\ref{eq_joint_pragmatic_listener}). This defines a probability for every possible reward function (e.g. every reward weight vector $w$). We then initialize our individual learner as described above.

When the individual learner performs Thompson sampling, it now performs an additional importance sampling step. Rather than sample a \emph{single} reward vector from its Gaussian prior, it samples a minimum of 100 possible reward vectors. As described above, it first discretizes these vectors then re-weights them according to the probability of each vector from its pragmatic social inference. Finally, it samples a single reward vector from this re-weighted set and uses this to choose an action. 

\subsection{Integrating literal information}

We use a similar procedure to test individual learning with our literal listener.
For descriptive utterances, we use the listener's posterior over reward functions (Eq.~\ref{eq_literal_listener_belief_update}). However, because there are a handful of false utterances in the experimental data (e.g. ``Spotted is -1''), using a hard constraint breaks the importance-sampling procedure described above. 
We therefore instead use soft-conditioning by setting a very low likelihood on inconsistent worlds ($\epsilon = 1^{-10}$) instead of ruling them out entirely. We use this posterior for importance sampling as described above.

For instructions, we modify the action selection step. We set the listener's policy to take the instructed action if available. If the action is not available, then they follow the Thompson sampling procedure described above. This is the simplest and most ``obedient'' interpretation of instructions~\cite{milli2017should}. We find that it yields rapid learning early on, as the instruction guides exploration. However, 57\% of instructions designate sub-optimal actions (see Fig.~\ref{fig:utterance-heatmaps}A).
A literal listener instructed to take one of these (e.g. ``Take solid green mushrooms'') is forced to continue taking them \emph{even after} inferring spotted green mushrooms are likely worth more. This constraint on their policy eventually leads their regret to asymptote below the more flexible pragmatic learner (Fig.~\ref{fig:individual_learning}). As discussed in the main text and noted in prior work~\cite{milli2017should}, more flexible approaches to instruction-following could avoid this pitfall.

\subsection{Simulation details}
All simulations were run on consumer hardware (a MacBook Pro). 
For each of the 2772 utterances in our behavioral experiment, we ran 5 independent Thompson sampling simulations, each spanning 25 timesteps. We repeated this process for each of the pragmatic listener models (Known $H$, Latent $H$, $H=1$, and $H=4$) in our experiment, giving us 13860 Thompson sampling simulations each model. We then ran the same number of independent simulations for the ``Individual'' learner (which used only the Gaussian prior described in Appendix~\ref{sec_appdx_individual_learning_thompson_sampling}).


\end{document}